\documentclass[runningheads]{llncs}
\usepackage[T1]{fontenc}
\usepackage{graphicx}
\newcommand{\orcid}[1]{\href{https://orcid.org/#1}{\includegraphics[scale=0.02]{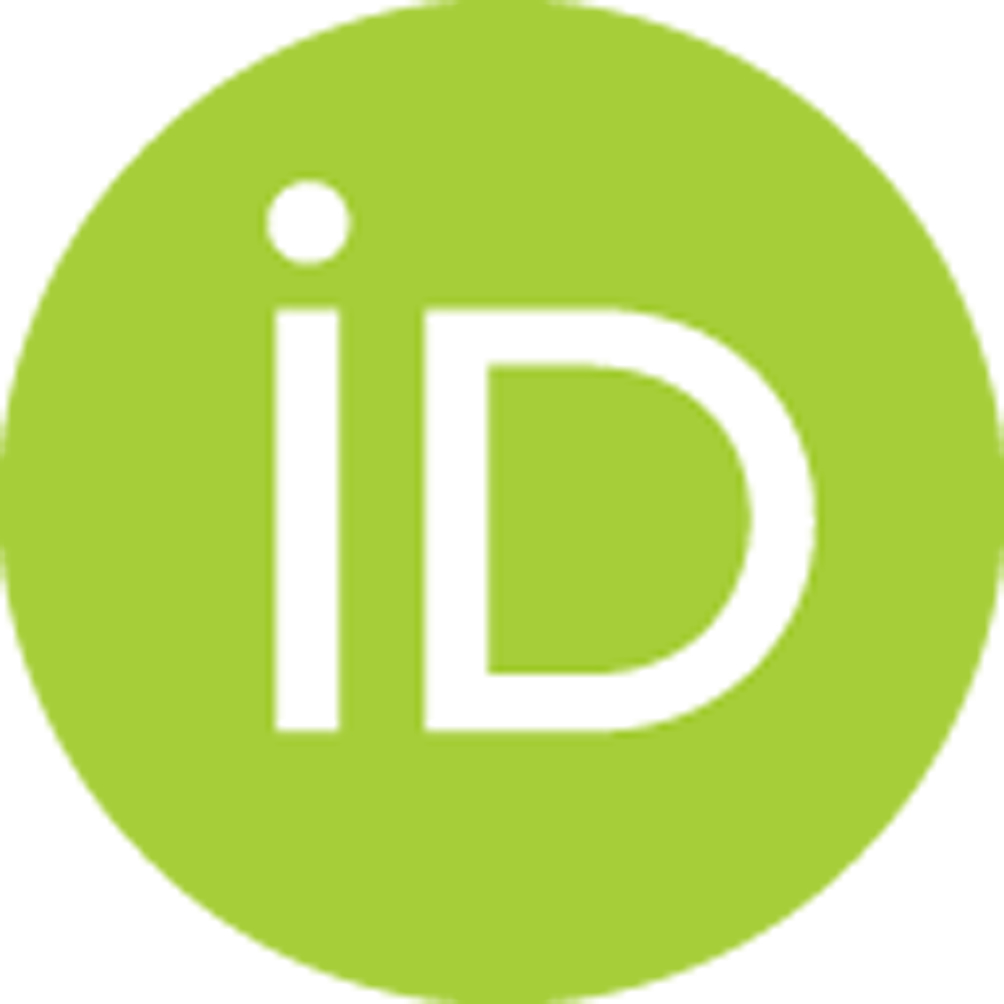}}}

\usepackage{amsmath}
\usepackage{amssymb} 
 \usepackage{hyperref}
\usepackage{booktabs}
\usepackage{geometry}

\begin{document}
\title{Classifying Histopathologic Glioblastoma Sub-regions with EfficientNet}
\author{%
  Sanyukta Adap\inst{1}\orcid{0009-0006-3639-0911}\and
  Ujjwal Baid\inst{1,2,3}\orcid{0000-0001-5246-2088}\and
  Spyridon Bakas\inst{1,2,3,4,5,6}\orcid{0000-0001-8734-6482}%
}
\institute{Division of Computational Pathology, Department of Pathology \& Laboratory Medicine, Indiana University School of Medicine, Indianapolis, IN, USA \and
Indiana University Melvin and Bren Simon Comprehensive Cancer Center, Indianapolis, IN, USA \and
Department of Radiology \& Imaging Sciences, Indiana University School of Medicine, Indianapolis, IN, USA \and
Department of Biostatistics \& Health Data Science, Indiana University School of Medicine, Indianapolis, IN, USA \and
Department of Neurological Surgery, Indiana University School of Medicine, Indianapolis, IN, USA \and
Department of Computer Science, Luddy School of Informatics, Computing, and Engineering, Indiana University, Indianapolis, IN, USA \and
\textdagger Corresponding authors: sadap@iu.edu, ubaid@iu.edu, spbakas@iu.edu\\
* Equal senior authors}

\authorrunning{S. Adap et al.}

\maketitle
\begin{abstract}
Glioblastoma (GBM) is the most common aggressive, fast-growing brain tumor, with a grim prognosis. Despite clinical diagnostic advancements, there have not been any substantial improvements to patient prognosis. Histopathological assessment of excised tumors is the first line of clinical diagnostic routine. We hypothesize that automated, robust, and accurate identification of distinct histological sub-regions within GBM could contribute to morphologically understanding this disease at scale. In this study, we designed a four-step deep learning approach to classify six (6) histopathological regions and quantitatively evaluated it on the BraTS-Path 2024 challenge dataset, which includes digitized Hematoxylin \& Eosin (H\&E) stained GBM tissue sections annotated for six distinct regions. We used the challenge's publicly available training dataset to develop and evaluate the effectiveness of several variants of EfficientNet architectures (i.e., B0, B1, B2, B3, B4). EfficientNet-B1 and EfficientNet-B4 achieved the best performance, achieving an F1 score of 0.98 in a 5-fold cross-validation configuration using the BraTS-Path training set. The quantitative performance evaluation of our proposed approach with EfficientNet-B1 on the BraTS-Path hold-out validation data and the final hidden testing data yielded F1 scores of 0.546 and 0.517, respectively, for the associated 6-class classification task. The difference in the performance on training, validation, and testing data highlights the challenge of developing models that generalize well to new data, which is crucial for clinical applications. The source code of the proposed approach can be found at the GitHub repository of Indiana University Division of Computational Pathology: \url{https://github.com/IUCompPath/brats-path-2024-enet}.

\end{abstract}
\section{Introduction}
Glioblastoma (GBM) is a fast-growing brain tumor affecting the functional brain tissue and its surroundings. Prognostically, the median survival of GBM patients is about 12-18 months post-diagnosis and following standard-of-care treatment; however, early diagnosis can significantly impact the patient survival outcomes~\cite{ref3}. A significant proportion of GBMs have an unknown cause. Some risk factors include previously undergone radiation therapy and hereditary conditions like neurofibromatosis and Li-Fraumeni syndrome. Typically, diagnosis is made using MRI or CT scans, followed by the histopathological assessment of a tissue biopsy~\cite{ref7}. GBMs have a diverse genetic profile with heterogeneous microenvironmental histopathologic characteristics and a highly infiltrative nature. Accurately identifying these tumors and evaluating their heterogeneity is crucial for determining the best course of action and potentially improving patient survival rates~\cite{ref4}.

Whole Slide Imaging (WSI) has become increasingly important in addressing this challenge of accurately detecting tumors, assessing their diverse histopathological features, and offering deeper insights into the tumor microenvironment. WSIs are digitized tissue sections, enabling the detailed examination of distinct morpho-pathological characteristics across large, high-resolution images. This technology, paired with artificial intelligence (AI) analysis, can contribute to furthering our understanding of this disease and provide information for more accurately diagnosing GBM, and hence guiding appropriate treatment strategies. 
With recent advancements in AI and data availability~\cite{ref9,ref10,ref11,ref12,ref13,ref20,ref21,ref4,ref24}, it has become easier to train AI models to automate the identification and classification of GBM~\cite{ref17,ref18,ref19}. This study proposes a four-step deep learning approach for classifying six histopathological regions in GBM using variants of the EfficientNet architecture. The presented method was designed and submitted for evaluation as part of the Brain Tumor Segmentation - Pathology (BraTS-Path) Challenge~\cite{ref4}. By addressing model complexity and overfitting, EfficientNet aims to create a reliable model for more precise diagnosis and personalized treatment planning for GBM patients.


\section{Materials and Methods}

\begin{table}[b]
\caption{BraTS-Path Training Set Distribution.}\label{tab1}
\centering
\begin{tabular}{|l|l|l|}
\hline
{\bfseries Class} &  {\bfseries Count} &  {\bfseries \%} \\
\hline
Cellular Tumor (CT) & 34139 & 35\%\\
Geographic Necrosis (NC) & 29542 & 31\%\\
Infiltration into Cortex (IC) & 14500 & 15\%\\
Psuedopalisading Necrosis (PN) & 9664 & 10\%\\
Microvascular Proliferation (MP) & 4812 & 5\%\\
Penetration into White Matter (WM) & 3828 & 4\%\\
\hline
{\bfseries Total} &  {\bfseries 96,485} &  {\bfseries 100\%} \\
\hline
\end{tabular}
\end{table}

\subsection{Data}
The BraTS-Path challenge was hosted on the Synapse Platform~\cite{ref22}. The participants were provided with datasets in 2 phases: training and validation. The labels were provided for the training data but not for the validation data. During the validation phase, the participants were asked to submit their predictions on validation data in a CSV file on the Synapse evaluation platform. Finally, during the testing phase, participants were asked to submit an MLCube\cite{ref25} container image of their model inference pipeline for evaluation on the hidden test dataset.

\begin{figure}[t]
\includegraphics[width=\textwidth]{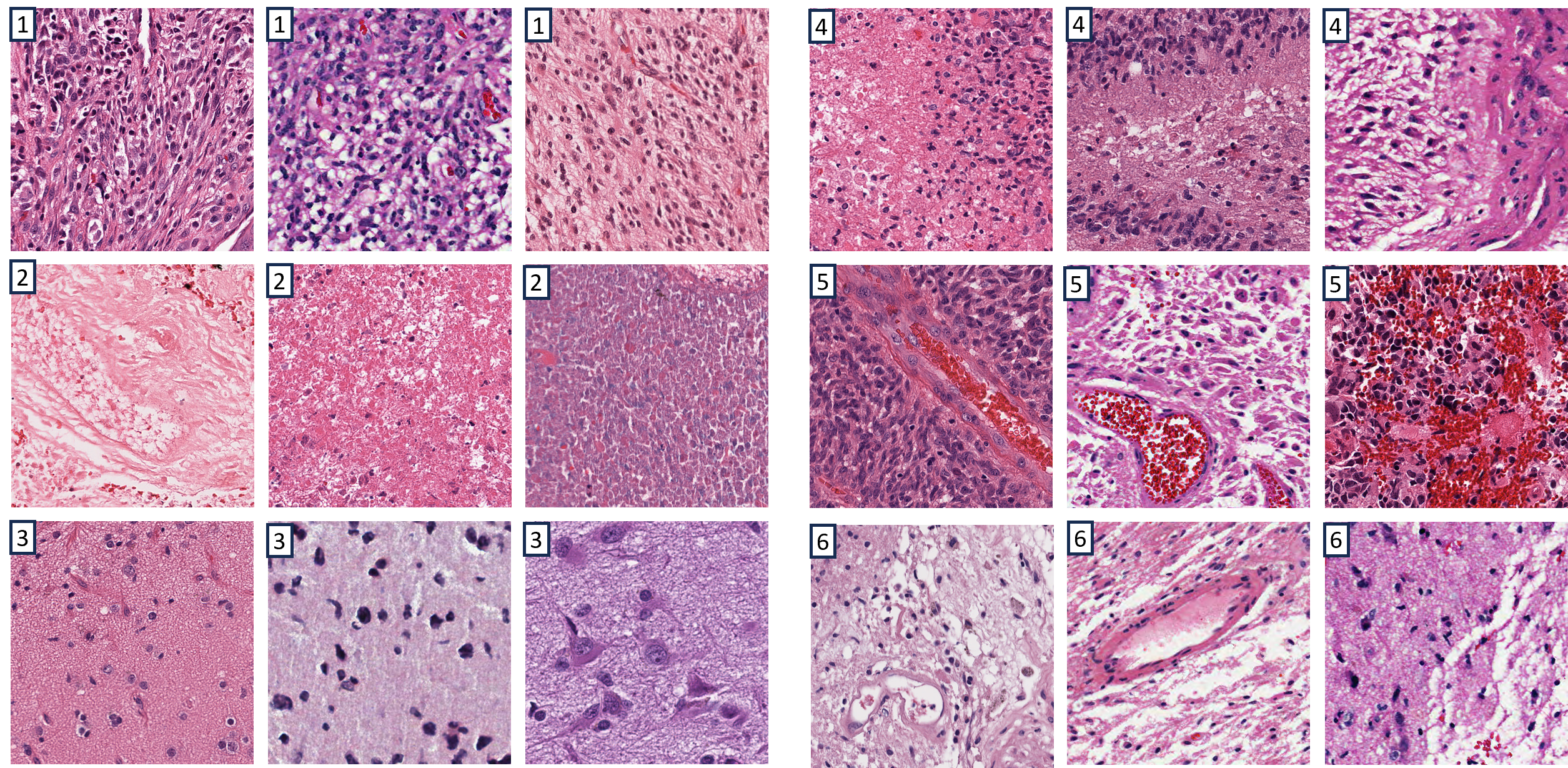}
\caption{Sample images in the training data. 1-Cellular Tumor (CT), 2-Pseudopalisading Necrosis (PN), 3-Infiltration into Cortex (IC), 4-Geographic Necrosis (NC), 5-Microvascular Proliferation (MP), and 6-Penetration into White Matter (WM).} 
\label{fig1}
\end{figure}

The training dataset consisted of 96,485 images divided into 6 distinct histological regions. The images were in PNG format and $512\times512$ pixels with 3 channels (RGB). The pixel values ranged from 0 to 255. The data distribution for each class can be found in Table~\ref{tab1}, along with some sample representative images of each class in Figure~\ref{fig1}. The challenge validation dataset consisted of 36,401 images of the same size and shape whose labels were not provided to the participants.

We divided the challenge's training dataset into our internal training and validation cohorts using an 80:20 split, each of them proportional to the given classes.

\subsection{Pre-processing}
Ensuring the input data is clean, consistent, and representative of the task allows the CNN to learn efficiently and generalize effectively to new data. This crucial step is fundamental for achieving precise and dependable outcomes in any CNN-based application. For this, we pre-processed data in three steps: 1) Deconvolution, 2) Data Augmentation, and 3) Normalization, to ensure smooth and efficient training. 

\subsubsection{Deconvolution:}
Deconvolution is a process of reading input images to RGB and converting them to PyTorch tensors. The images were transformed into numerical arrays that represented the pixel values from the original PNG image. To make the pixel intensities appropriate for AI methods, they must be converted into multi-dimensional arrays or tensors. For gray-scale images, every pixel value within these arrays represents its intensity; for color images, it represents its RGB components.

\subsubsection{Data Augmentation:}

During this pre-processing stage, images are augmented to enhance the dataset's variability by implementing a series of transformations. These include random rotations up to 20 degrees, random horizontal and vertical flips, as well as random modifications to brightness and contrast by 0.1. These augmentations generate diversity by creating different variations of each image, thus increasing the number of unique data points with every iteration through the dataset.

\subsubsection{Normalization:}
Further, the pixel values of each image were normalized to a floating-point type and scaled to center around 0. The scaling was done by first calculating the mean and standard deviation for all the images in the mini-batch. Then, the images were normalized using (1): 

\begin{equation}
Z_i = \frac{X_i - \mu}{\sigma + \epsilon}
\end{equation}

Where, \( \mu \) is the global mean of the image set \( X \), \( \sigma \) is the standard deviation, \( \epsilon = 1 \times 10^{-10} \) is a very small constant added to the denominator which prevents the denominator from turning zero, \( i = [1 - 2083] \) is the index of each training sample, and \( Z_i \) is the normalized version of \( X_i \). Normalization ensures that the inputs in the batch are on a similar scale. Normalizing/standardizing inputs has several benefits:

\begin{enumerate}
    \item prevents domination of specific features with higher pixel values, in the learning process. 
    \item balances the gradients that direct the model learning, lowering the likelihood of oscillations or slow convergence.
    \item minimizing instabilities due to larger numerical values, leading to more consistent results across different batches.
    \item by standardizing the scale of input images, the model can better understand the underlying patterns in the data instead of learning specific pixel value ranges.
\end{enumerate}

\subsection{Model Training}
\subsubsection{EfficientNet:}
EfficientNets~\cite{ref6} (ENets) are a range of models that have made model scaling more systematic by balancing the depth (number of layers), width (number of channels), and input image resolution of CNN models using predefined scaling coefficients. This approach is referred to in the literature as compound scaling~\cite{ref6}. 

ENet model ranges from B0 to B7. The ENet-B0 is the smallest in the range and is considered a baseline model due to its mobile-sized architecture. The variants B1 to B7 are scaled-up versions of B0. The ENet models have achieved state-of-the-art performance on ImageNet~\cite{ref8} and other well-known transfer learning datasets in 2019 and 2020 that are widely used for training and evaluation of Convolutional Neural Networks (CNNs)~\cite{ref6}. 

In this study, we used transfer learning to train ENet-B0 to ENet-B4 pre-trained on ImageNet~\cite{ref8}. The last layer of the model was modified to have 6 nodes corresponding to 6 classes of the BraTS-Path data, and the training was conducted from scratch.

\subsubsection{Cross-Validation:} 
The training was done in a 5-fold cross-validation to evaluate the performance of models on new unseen data. For this, we divided the dataset into equal parts in a stratified manner to maintain the distribution of classes across the folds.

\subsubsection{Loss:}
After one epoch of training, the model predicts the output in the form of logits for each image in the mini-batch. These outputs were passed to a Weighted Cross Entropy Loss Function, L, (2) to compute the classification error across each mini-batch.

\begin{equation}
L = -\frac{1}{N} \sum_{i=1}^N w_{y_i} \cdot \log(\hat{y}_{i}, {y_i})
\end{equation}

Where: \textit{N} is the number of samples, \textit{\( w_{y_i} \)} is the weight for the true class \( y_i \) of the \( i \)-th sample, \textit{\(\hat{y}_{i}\)} is the predicted probability for the true class \( y_i \) of the \( i \)-th sample.

\subsubsection{Metrics:}
After each epoch, both training and validation (training and validation split from the original training data) metrics were calculated. We evaluated the models based on Loss (2) and F1 scores per epoch. We also plotted a confusion matrix for each epoch and saved the model weights in the form of checkpoints.

\subsection{Inference}
For inference, the final set of hyperparameters, shown in Table~\ref{hyperparams}, was used to train the top-performing models in a 5-fold cross-validation manner. We selected 5 models, one from each fold, and ensembled their logits by taking their average and then applying a softmax function over it. This way, we combine the results from each ``weak'' model to form one ``strong'' model, which can classify the images much better and robustly than each of the individual models, while avoiding systematic biases of the individual models.

\subsection{Experimental Design}

\begin{table}[t]
\caption{Experiment Hyperparameters.}
\label{hyperparams}
\centering
\begin{tabular}{|l|l|}
\hline
{\bfseries Hyperparameters} &  {\bfseries Value}\\
\hline
Batch Size & 16 \\
Number of Epochs & 60\\
Loss Function & Cross Entropy Loss \\
Optimizer & Adam \\
Momentum & 0.0001 \\
LR & 0.001 \\
LRS & Exponential \\
Gamma & 0.9 \\
\hline
\end{tabular}
\end{table}

We conducted all experiments on Indiana University's High-Performance Computing Cluster, BigRed200, by requesting 256 GB of memory, a single 64-core, 2.0 GHz, 225-watt AMD EPYC 7713 processor, and four NVIDIA A100 GPUs. The codebase of this study is exclusively built on Python/PyTorch~\cite{ref5} and can be found at the Indiana University Division of Computational Pathology's GitHub repository: https://github.com/IUCompPath/brats-path-2024-enet.

We have referred to the validation data provided by the challenge as ``Challenge Validation Data'' since the labels were not provided to the participants. All models were initially evaluated on our internal validation split (created from the labeled BraTS-Path training dataset provided by the challenge organizers), and the final quantitative performance evaluation of our proposed approach was conducted on the challenge validation data. We carried out the following experiments to identify the optimal model.

First, the dataset undergoes the preprocessing steps in mini-batches of size 16. These mini-batches are small random subsets (or batches) of the dataset that pass through the model and update the parameters after one pass through the images in the mini-batch. This helps the model learn faster since it does not have to wait until the end of the dataset to calculate the gradients. 

Further, for the initial screening, we trained the models without cross-validation to identify which ENet models performed better on the validation data provided by the challenge. We also experimented with Learning Rate Schedulers (LRS)\cite{ref14} along with different rates of LR Decay (gamma). LRS is a method that tweaks the LR by a factor during training. This allows for a staged fine-tuning of the LR by enabling the model to make larger updates early in the training process when the parameters are far from optimal and smaller updates later on when the parameters are closer to the minima. For the experiments, we used 3 LRS provided by PyTorch:

\begin{enumerate}
    \item Step LR: This decays the LR by a factor of gamma at every predetermined period of LR decay.
    \item Multi-step LR: This decays the LR by a factor of gamma at a predefined epoch number.
    \item Exponential LR: This decays the learning rate by a factor of gamma at every epoch.
\end{enumerate}

All the models were trained for 60 epochs. For optimization, we used Adam~\cite{ref14} with a momentum of 0.0001. We performed multiple rounds of training in combination with different hyperparameters to find the models that had the best performance on the internal validation split. 

Then, with the top 2 models that performed best on the challenge validation data, we trained the models in a 5-fold cross-validation manner with the best combination of hyperparameters. For each fold, the input images, after pre-processing, were passed to the models in mini-batches of size 16. The model performs feature extraction, followed by batch normalization and adaptive average pooling. Finally, the images were passed to the classification layer of size 6 (number of classes).

Once all models had their respective predictions, we combined their results by averaging the outputs from each fold. This averaged output is the final prediction, which was submitted for evaluation on the challenge validation data.

\section{Results}
\subsection{Model Training}

The initial training without cross-validation revealed that ENet-B1 and ENet-B4 were the top 2 models among the 5, with a challenge validation F1 of 0.462 and 0.43, respectively. Table~\ref{tab3} presents the results from the initial experiments on the ENet Models (using hyperparameters mentioned in Table~\ref{hyperparams}) along with their performance metrics on the challenge validation dataset.

\begin{table}[t]
\caption{Initial Training Experiments}\label{tab3}
\centering
\begin{tabular}{|c|c|c|}
\toprule
Model & Challenge Validation F1 & Challenge Validation MCC\\
\midrule
ENet-B0 & 0.288 & 0.242 \\
\textbf{ENet-B1} & \textbf{0.462} & \textbf{0.374}\\
ENet-B2 & 0.345 & 0.262 \\
ENet-B3 & 0.274 & 0.214 \\
\textbf{ENet-B4} & \textbf{0.43} & \textbf{0.336}\\
\bottomrule
\end{tabular}
\end{table}

\begin{figure}[t]
\includegraphics[width=\textwidth]{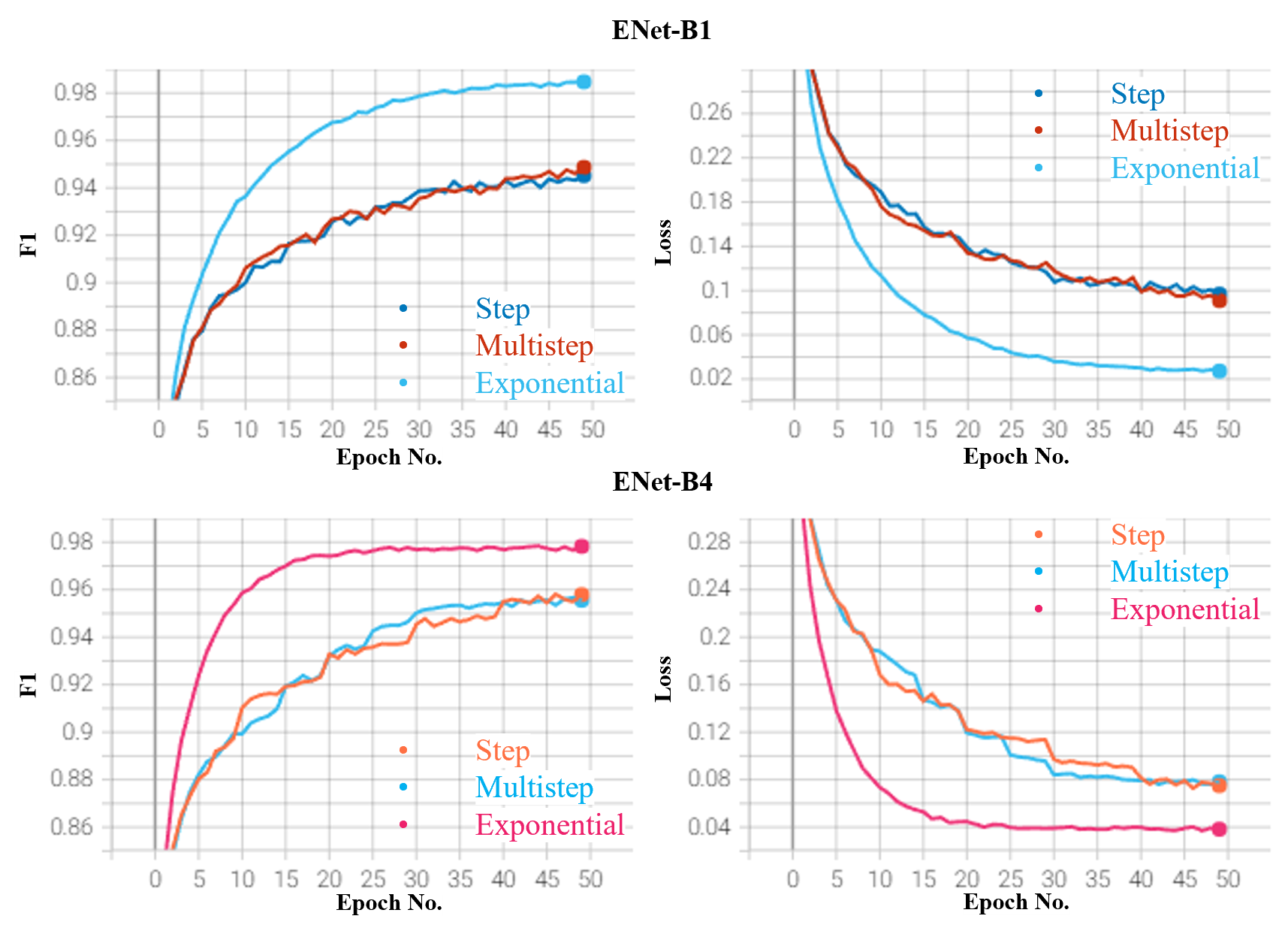}
\caption{Training F1 scores and Loss Curves for ENet-B1 and ENet-B4 per Epoch with Gamma = 0.9.} 
\label{fig2}
\end{figure}

Further experiments with LRS show that exponential LRS with a gamma of 0.9 yielded the best performance on the internal validation split. See Figure~\ref{fig2} for training curves.

\begin{figure}[t]
\includegraphics[width=\textwidth]{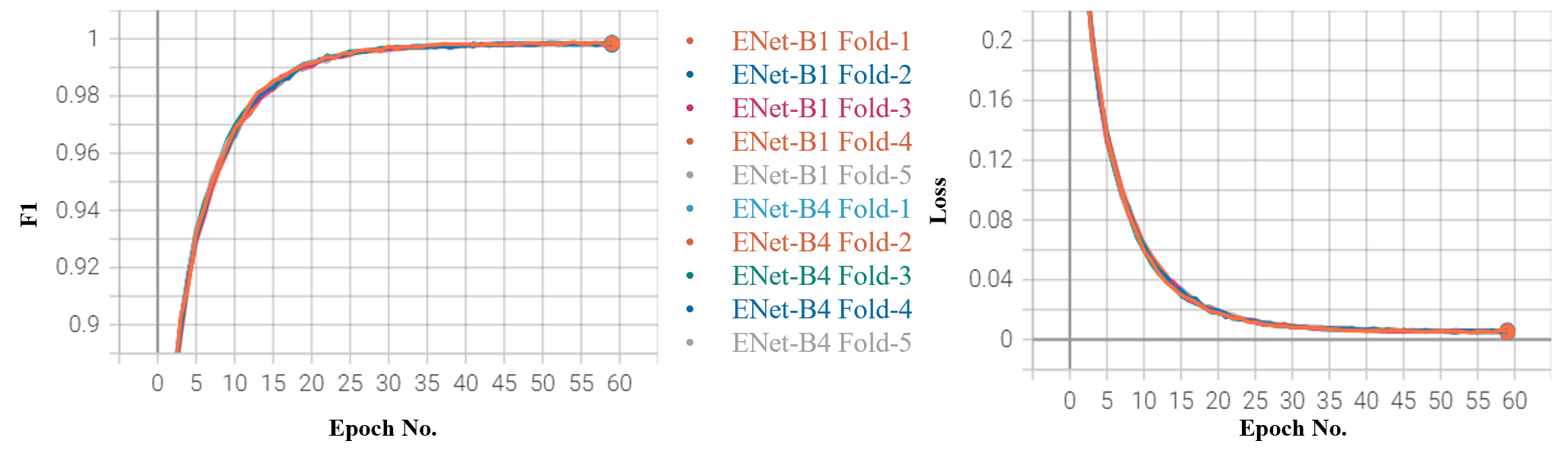}
\caption{Cross Validation Training Metric Curves for ENet-B1 and ENet-B4 per Epoch.} 
\label{curves}
\end{figure}

Next, with the final set of hyperparameters, we trained ENet-B1 and ENet-B4 in a 5-fold cross-validation manner. The performance of both models across all folds during the complete training was almost identical (Figure~\ref{curves}). Specifically, the convergence curves relative to both the performance evaluation metric (F1) and loss are overlapping.

\subsection{Inference}
Once the models were trained, we chose model checkpoints across all folds at epochs 20, 25, 30, and 35. Then, for each ENet model, we chose a model checkpoint for each fold, and combined/ensembled their predictions on the challenge validation data. The results of both models with model checkpoints at different epochs are shown in Table~\ref{tab5}.

\begin{table}[b]
\caption{Challenge Validation Set Metrics by Ensembling the predictions across folds.}
\label{tab5}
\centering
\begin{tabular}{|c|c|c|c|}
\toprule
Model & Model Checkpoint & Challenge Validation F1 & Challenge Validation MCC \\
\midrule
ENet-B1 & 25 & 0.491 & 0.388 \\
\textbf{ENet-B1} & \textbf{30} & \textbf{0.546} & \textbf{0.436} \\
ENet-B1 & 35 & 0.53 & 0.421 \\
ENet-B4 & 20 & 0.403 & 0.311 \\
\textbf{ENet-B4} & \textbf{25} & \textbf{0.438} & \textbf{0.337} \\
ENet-B4 & 30 & 0.39 & 0.301 \\
\bottomrule
\end{tabular}
\end{table}

The best-performing model, in this case, the EfficientNet-B1, was submitted for evaluation on the testing data in the testing phase. The results revealed an F1 score of 0.517 and an MCC of 0.380.

\section{Discussion}
In this study, we presented a 4-step deep learning approach with an EfficientNet model that can robustly classify 6 histologically distinct GBM regions. We mainly focused on training ENet models with the BraTS-Path training data and evaluated them using the BraTS-Path validation data. According to the results obtained from various experiments, the performance metrics of ENet-B1 seem to be higher than those of ENet-B4. This could be due to the larger and more complex architecture of ENet-B4 compared to ENet-B1.

Beyond ENets, during the initial screening of models, along with ENet variants, we also evaluated the ResNet~\cite{ref15} and DenseNet~\cite{ref16} variants. However, these models demonstrated significantly lower performance compared to the ENet models, which led to the decision of excluding their results from this manuscript. The disparity in performance can be explained by several key factors. While ResNet and DenseNet architectures are undeniably powerful, they lack the level of architectural efficiency and optimization found in ENet. ENet's compound scaling approach allows it to achieve higher performance with fewer parameters. Moreover, ENet leverages advanced design techniques like Mobile Inverted Bottleneck Convolutions (MBConv) and depthwise separable convolutions, which are both computationally efficient and improve the model's generalization capabilities. In contrast, the lower performance of ResNet and DenseNet in our evaluations can likely be attributed to their less optimized use of computational resources and parameters, despite their complex and deep designs. As a result, their outcomes did not warrant inclusion in the final analysis.

Further experimentation with LRS revealed that exponential LRS with gamma equal to 0.9 had the best performance. This could be likely due to a smooth and gradual decay of LR over the epochs. Since exponential LRS reduces LR exponentially, we used an LR of 0.001, which is a slightly higher value if used without LRSs. However, an exponential LRS ensured that the LR remained sufficiently high during the early epochs to learn quickly and gradually reduced it as the model approached better parameter estimates, leading to improved validation performance over time.

The small difference between the training and validation metrics can be attributed to the fact that the validation set was derived from the same dataset obtained from the challenge training website. This dataset was split into 80\% for training and 20\% for validation, meaning the validation set was ``in-sample'' and not truly independent. As a result, the model's performance on this validation set was high (0.95+). In contrast, the challenge validation and testing set likely had a different, ``out-of-sample'' distribution, leading to lower scores when evaluated on that cohort.

Overall, ENet-B1 and ENet-B4 training both followed a fairly similar trend. The models' performance on the training set was comparable, reflected by the overlapping training loss and F1 curves shown in Figure~\ref{fig1}, suggesting that both models could learn from the training data effectively and without fluctuations throughout the training process. Furthermore, it implies that both models are proficient in identifying the fundamental characteristics of the training data from BraTS-Path. The simpler ENet-B1 generalizes better to challenge data than the more complicated ENet-B4, but the difference in performance on the challenge validation and testing data emphasizes the significance of model complexity and the possibility of overfitting. This result highlights how important it is to maintain a balance in machine learning tasks between model complexity and its ability to generalize the data.

The observed F1 scores on validation and testing sets indicate that the model’s predictions are not random. In a six-class classification task, random predictions would yield an F1 score of around 0.172 ($\frac{1}{6}$). Our model achieved substantially higher scores of 0.546 and 0.517 on validation and testing cohorts, respectively, demonstrating its ability to meaningfully differentiate histopathological sub-regions. This suggests that the model is learning relevant morphological patterns, underscoring its potential in clinical applications. However, the performance gap between training and holdout data highlights the need for further refinement to improve generalizability.

The implications of our study are significant for the field of computational pathology and the diagnosis of GBM. It suggests that simpler models like ENet-B1 may generalize better to new data when compared to more complex models, which can overfit despite their larger capacity. This highlights the need to balance model complexity with generalization, especially in medical applications where accuracy is crucial. Additionally, the study underscores the importance of diverse training datasets to improve the performance on real-world cases, potentially leading to more reliable AI-driven diagnostic tools and better patient outcomes.

Our contribution demonstrates that a streamlined model like ENet-B1 can achieve superior generalization in GBM classification compared to more complex architectures. This challenges the conventional focus on larger models and suggests that efficiency and simplicity may often lead to better performance in certain medical contexts. Additionally, our study underscores the importance of dataset diversity in developing models that can be trusted in real-world clinical applications.

The study advances our understanding of GBM classification by demonstrating that simpler models like ENet-B1 can generalize effectively to unseen data, a crucial factor in accurately diagnosing brain tumors. It reveals that complex models, prone to overfitting, may not always be ideal for biological data where variability is high. This insight highlights that efficiency and simplicity in model architecture can enhance performance in medical applications by better capturing the essential biological features of GBM. Furthermore, the research emphasizes the necessity of diverse and representative training datasets to reflect the biological diversity of real-world cases, ultimately improving the robustness and clinical applicability of AI-driven diagnostic tools.

Despite the results, this study has several limitations. Firstly, the BraTS-Path training dataset may not accurately reflect the diversity of GBM cases seen in clinical settings. Furthermore, the BraTS-Path multi-institutional training data was split into training and validation cohorts, potentially resulting in an ``in-sample'' situation that would have inflated the validation performance metrics. Since the challenge validation set and the labeled validation split appear to come from completely different distributions, this setup does not mirror real-world conditions where the validation set should ideally represent the test data. In practice, the validation set serves as a proxy for unseen test data, and it should reflect the characteristics of that test data to provide a meaningful estimate of how the model will generalize. If the validation set is too similar to the training set, it may lead to inflated performance estimates and failure to accurately assess model robustness. Moreover, we did not investigate the effects of regularization methods such as data augmentation, which may reduce overfitting in larger models. Lastly, we focused on one particular model, the ENet, without exploring other architectures that might offer a better trade-off between generalization and complexity. While several state-of-the-art models may have been better suited for this task, the strict timeline of the challenge and our limited computational resources necessitated the exclusive use of ENet models. This choice allowed us to strike a balance between performance and efficiency, helping us meet the challenge's deadlines while making the most of the resources available.

Future directions might focus on mitigating these limitations by extending the dataset to encompass more diverse sources of data, thereby better representing the heterogeneity of GBM cases observed in clinical settings. Furthermore, the reliability of model evaluation may be increased by using cross-validation procedures, but with carefully chosen validation sets that more closely resemble the real-world test data. Additionally, experimenting with various current state-of-the-art architectures, using regularization strategies (data augmentation and dropout), and utilizing a different optimizer may improve the robustness and generalization of the model. Future research may be able to solve these issues to provide more promising outcomes.

With this study, we aimed to develop a deep learning model that can accurately classify histopathological regions in GBM tissue sections, with the ultimate objective of improving diagnostic accuracy and facilitating quantitative studies towards furthering our disease understanding. By leveraging the ENet architecture and utilizing the BraTS-Path challenge dataset~\cite{ref4}, our goal was to create a robust and generalizable model capable of distinguishing critical histopathological features of GBM.

\section*{Acknowledgments}
This research was supported in part by Lilly Endowment, Inc., through its support for the Indiana University Pervasive Technology Institute.




%
%
%
%

\end{document}